\documentclass{article}
\usepackage{spconf,amsmath,graphicx}
\usepackage{url}
\usepackage{lipsum}  
\usepackage{booktabs}
\usepackage{todonotes}

\renewcommand\vec{\mathbf}

\ninept
\title{Transformer-based encoder$\cdot$encoder architecture\\for Spoken Term Detection}
\name{Jan Švec, Luboš Šmídl, Jan Lehečka}
\address{\vspace{-1mm}Department of Cybernetics, University of West Bohemia, Pilsen, Czech Republic\\\small\texttt{[honzas,smidl,lehecka]@kky.zcu.cz}}
\begin{document}
%
\maketitle
\begin{abstract}
The paper presents a method for spoken term detection based on the Transformer architecture. We propose the encoder$\cdot$encoder architecture employing two BERT-like encoders with additional modifications, including convolutional and upsampling layers, attention masking, and shared parameters. The encoders project a recognized hypothesis and a searched term into a shared embedding space, where the score of the putative hit is computed using the calibrated dot product. In the experiments, we used the Wav2Vec 2.0 speech recognizer, and the proposed system outperformed a baseline method based on deep LSTMs on the English and Czech STD datasets based on USC Shoah Foundation Visual History Archive (MALACH). 
\end{abstract}
\section{Introduction}
\label{sec:intro}

Searching through large amounts of audio data is a common feature of several tasks in speech processing, namely keyword spotting (KWS), wake word detection (WWD), query-by-example (QbE), or spoken term detection (STD). These tasks differ both with the requirements imposed on the form of the query: for example, audio snippet (QbE) or sequence of graphemes (STD) and computational resources required: low resource (WWD), real-time processing (KWS), or off-line processing (STD). In recent works, the architecture of such systems is often based on acoustic embeddings extracted using deep neural networks \cite{settle2016awe}. Such embeddings are further used to classify or detect keywords, terms, or examples. The embeddings could be extracted using convolution networks, recurrent networks, or -- more recently -- Transformer networks.

Using Transformer architecture for the mentioned tasks can be seen in many recent papers. For example, the Keyword Transformer~\cite{berg21_interspeech} is designed for the Google Speech Command (GSC) task, which directly uses the Transformer architecture to project the input audio data into a single vector used for keyword classification. Another example of a similar approach is the architecture called LETR \cite{ding2022_letr}. Unfortunately, the GSC-related works often present a solution for keyword classification instead of KWS or STD in large, streamed audio data.

The multi-headed self-attention mechanism is also used in the QbE scenario. For example, \cite{yuan2022} proposes a combination of recursive neural networks, self-attention layers, and a hashing layer for learning binary embeddings used for fast QbE speech search.

The works related to WWD also study the use of the Transformer architecture. The streaming variant of Transformer was proposed in \cite{wang2021}. The Transformer was modified for real-time usage and outperformed the system based on the convolution network. Another architecture called Catt-KWS \cite{yang22n_interspeech} uses a cascaded Transformer in the encoder-decoder setup for real-time keyword spotting.

In this work, we do not focus on the direct processing of the input speech signal. Instead, we use the speech recognizer to convert an audio signal into a graphemic recognition hypothesis. The representation of speech at the grapheme level allows preprocessing the input audio into a compact confusion network and further to a sequence of embedding vectors. In \cite{svec21_interspeech}, we proposed a Deep LSTM architecture for spoken term detection, which uses the projection of both the input speech and searched term into a shared embedding space. The hybrid DNN-HMM speech recognizer produced phoneme confusion networks representing the input speech. The DNN-HMM speech recognizer can be replaced with the Wav2Vec 2.0 recognizer \cite{Baevski2020} with CTC loss -- an algorithm for converting the CTC grapheme posteriors into grapheme confusion network was proposed in \cite{svec22_interspeech}. The grapheme confusion networks were subsequently used in the Deep LSTM STD. Moving from DNN-HMM to the Wav2Vec speech recognizer significantly improved the STD performance. 

This work describes a modification of the STD neural network (Sec. \ref{sec:neural}) by replacing the Deep LSTMs with the Transformer encoder (Sec. \ref{sec:transformer}). The core of the Transformer encoder has the same architecture as BERT-like (Bidirectional Transformers for Language Understanding) models \cite{devlin2018bert, Liu2019b} (Sec. \ref{sec:block}), but a simple drop-in replacement of LSTMs with vanilla Transformer encoder brings a significant degradation in STD performance (see Tab. \ref{tab:results}) which contradicts the common understanding of Transformers as the more superior class of models. This interesting observation motivated the research presented in this paper. To overcome the LSTM baseline, we propose a set of modifications (Sec. \ref{sec:proposed}). The proposed method is experimentally evaluated in the domain of oral history archives (Holocaust testimonies from the USC Shoah Foundation Visual History Archive) in two languages (Sec. \ref{sec:dataset}). The experimental results (Sec. \ref{sec:experimental}) show that the proposed Transformer architecture outperforms the baseline Deep LSTM.

\section{Neural networks for spoken term detection}
\label{sec:neural}

Our design of STD based on neural networks consists of two independent processing pipelines: (1) the recognition output projection by a \emph{hypothesis encoder} and (2) the searched term projection and minimum length estimation by a \emph{query encoder}. Suppose the input audio is represented by the recognized hypothesis, which is a sequence of time-aligned segments $c_i,~i=1, \ldots N$. Each segment $c_i$ is projected into a vector $\vec{C}_i$. The query $g$ is expressed as the sequence of graphemes $g_j,~j=1, \ldots M$ mapped using an input embedding layer to vectors $\vec{G}_j$. 

The hypothesis encoder is used to map the sequence of vectors $\vec{C}_{i=1}^N$ to a sequence of embedding vectors $\vec{R}_{i=1}^N$. The query encoder maps the vectors $\vec{G}_{j=1}^M$ to a sequence of query embeddings $\vec{Q}_{k=1}^K$. Here we assume that the length of sequences $\vec{C}_i$ and $\vec{R}_i$ is the same to keep the time correspondence of $\vec{R}_i$ to the input audio. It is not necessary for the query and generally $M \neq K$. For example, the query can be represented by a single vector ($K=1$) or, like in \cite{svec21_interspeech}, by three vectors ($K=3$, in this case $\vec{Q}_1$ represents the first half of the query, $\vec{Q}_2$ the middle of the query, and $\vec{Q}_3$ the second half).

\begin{figure}[t]
\centering
\centerline{\includegraphics[width=0.5\textwidth]{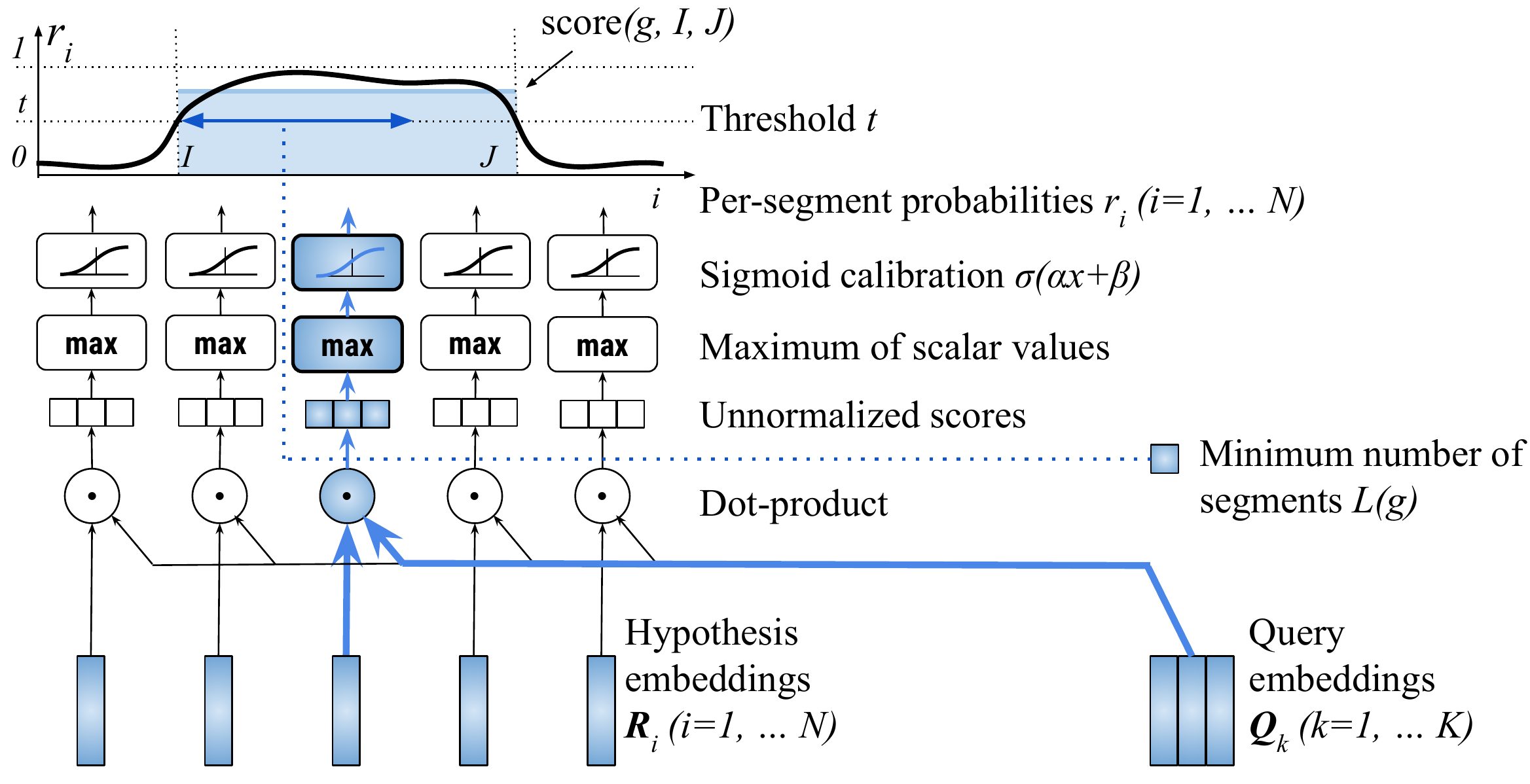}}

\caption{Neural networks for STD. Per-segment probabilities are intentionally displayed as a continuous function for clarity, although they are represented as discrete values $r_i$.}
\label{fig:nn-based}
\vspace{-1em}
\end{figure}

The embedding vectors $\vec{R}_i$ and $\vec{Q}_k$ are then used to compute per-segment probabilities of segment $c_i$ being the part of the putative hit of the query $g$. To compute the calibrated probabilities $r_i, i=1, \ldots N$, we use the dot-product of the embedding vectors:

\begin{equation}
    r_i = \sigma\left(\alpha \cdot \max_{k=1}^K ( \vec{R}_i \cdot \vec{Q}_k ) + \beta\right)
    \label{eq:prob}
\end{equation}

\noindent where $\sigma(x)=\frac{1}{1+e^{-x}}$ denotes the sigmoid function and $\alpha$ and $\beta$ are trainable calibration parameters. The maximum is used to select the most similar (in terms of dot-product) query embedding $\vec{Q}_k$ of all $K$ embeddings (see Fig. \ref{fig:nn-based}).

To determine the putative hits of the query $g$, the minimum number of segments $L(g)$ is estimated in the query encoder and used to find all spans $(I, J)$ satisfying the conditions $r_i > t ~~\forall i: I \leq i \leq J$ and $J-I+1 \geq L(g)$ where $t$ is the decision threshold ($t=0.5$ is used in the experiments). In other words, we search for peaks in $r_i$ threshold $t$ which span at least $L(g)$ time-aligned segments $c_i$. The overall score for the putative hit is determined as an average probability:

\begin{equation}
    \mathrm{score}(g, I, J) = \frac{1}{J-I+1} \sum_{i=I}^J r_i
\end{equation}

\noindent Note, that if there are multiple overlapping spans for a given query, only the span with the highest score is kept as a putative hit.

The NN-based model is trained using a binary cross-entropy as a~loss function. The training data can be generated on the fly by randomly selecting a word from a time-aligned transcript as a query. Because we are targeting oral history archives, we can exploit a~huge amount of speech data they contain by blindly recognizing the speech data and using the correctly recognized in-vocabulary words (in terms of confidence scores) for generating the queries. Because the main focus of STD in oral history archives is on the out-of-vocabulary words, we can simulate them by merging two or more consecutive in-vocabulary words \cite{svec21_interspeech}.

\section{Transformer-based spoken term detection}
\label{sec:transformer}

In this section, we will describe an application of the Transformer neural network architecture to the STD task. The basic idea of our proposed architecture is to use just the encoder part of the Transformer to extract context-dependent vector representations of the input. This approach is similar to using a Transformer in the BERT family of models \cite{devlin2018bert}. The novelty of our approach is to convert both the input audio and the searched query into a shared embedding space and then score each segment of the input audio using a simple similarity measure consisting of a sigmoid-calibrated dot-product between two vectors in this embedding space. In an analogy to the encoder-decoder approach, we can call this architecture the \emph{encoder$\cdot$encoder} (encoder-dot-encoder).

The proposed STD system does not process the input audio directly. Instead, it uses the fine-tuned Wav2Vec 2.0 model to recognize the grapheme-based representation of the input converted into the grapheme confusion network using the procedure described in~\cite{svec22_interspeech}. The grapheme confusion network representation allows using the graphemes (and their alternatives) of the recognition hypothesis, the timing of each grapheme, and the corresponding posterior probability (Fig. \ref{fig:feature-extraction}). The query is represented directly as a sequence of query graphemes. 

\subsection{Proposed architecture}
\label{sec:proposed}

To improve the performance of the Transformer-based STD model, we suggest some extensions of the vanilla BERT-like architecture.

\begin{figure}[t]
\centering
\centerline{\includegraphics[width=0.37\textwidth]{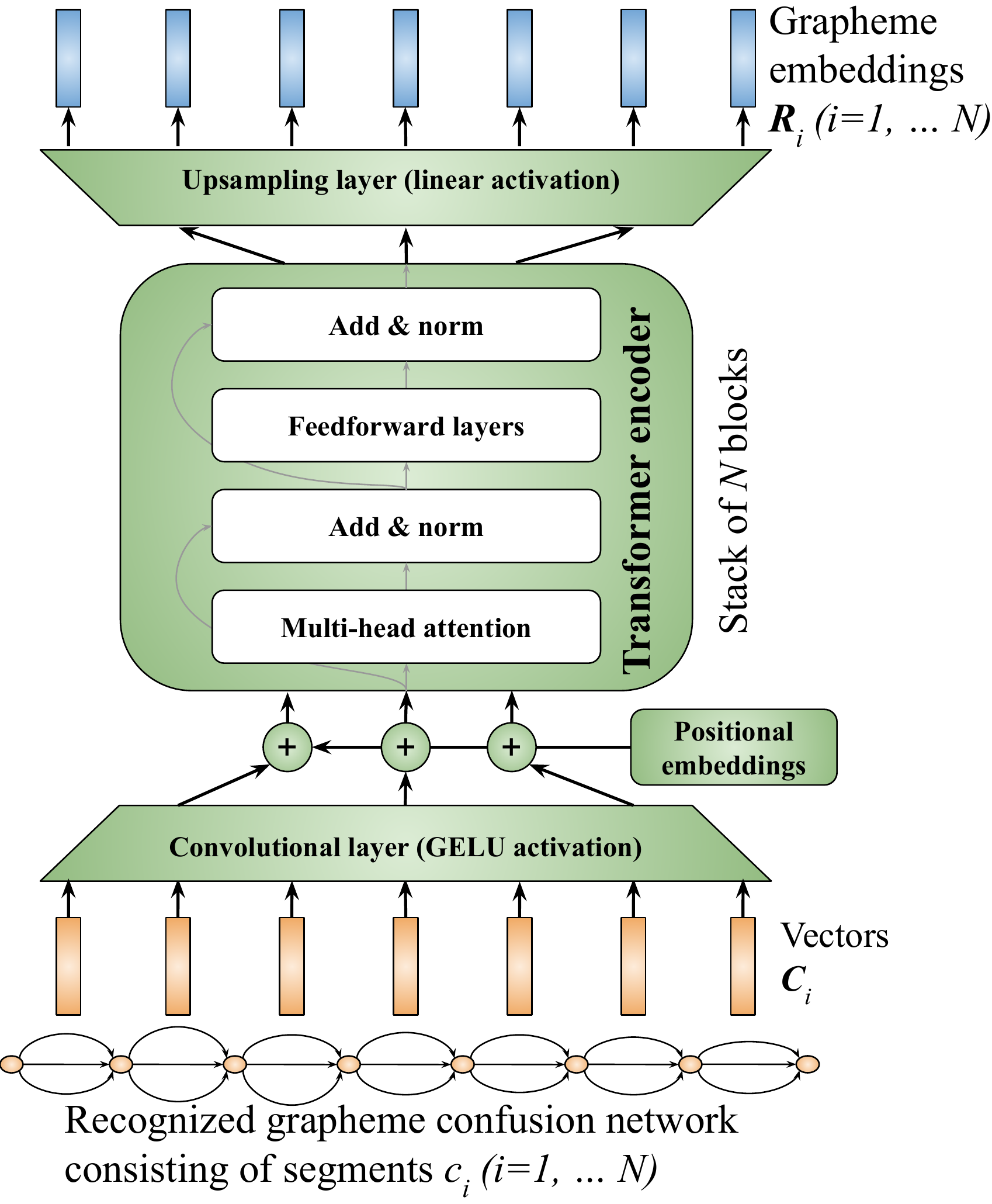}}
\vspace{-0.5em}
\caption{Architecture of the hypothesis encoder.}
\label{fig:hypothesis-encoder}
\vspace{1em}
\centerline{\includegraphics[width=0.4\textwidth]{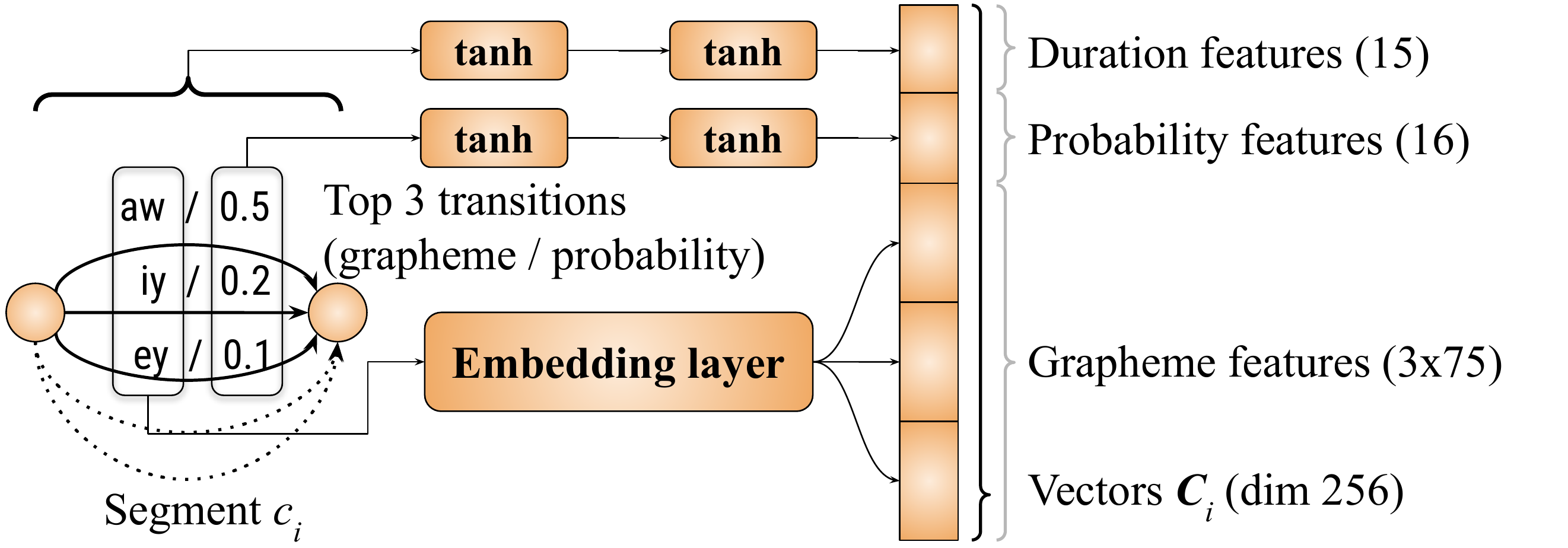}}
\vspace{-1em}
\caption{Mapping of confusion network segments $c_i$ to vectors $\vec{C}_i$.}
\label{fig:feature-extraction}
\vspace{-1em}
\end{figure}

First, we hypothesize that the worse performance of the vanilla Transformer architecture is caused by the format of the input -- the single graphemes. Therefore, we used the trainable \emph{1D convolutional filter} with a stride larger than 1 to reduce the temporal dimensionality of the input and to find projections of larger subsequences of graphemes. The Transformer is applied on top of such projections (output of the convolutional layer). To restore the input-output relation  with the time-alignment of the segments $c_i$, we also applied the upsampling layer (also called transposed convolution or deconvolution) on the output of the Transformer (Fig. \ref{fig:hypothesis-encoder}). BERT-like architectures \cite{devlin2018bert} use the GELU functions \cite{hendrycks2016gelu} as activations and therefore we used it not only in the encoder but also as activations in the convolutional layer. The upsampling layer employed a linear activation. The standard positional embeddings are added after applying convolution before feeding input into the Transformer.

\begin{figure}[t]
\centering
\centerline{\includegraphics[width=0.33\textwidth]{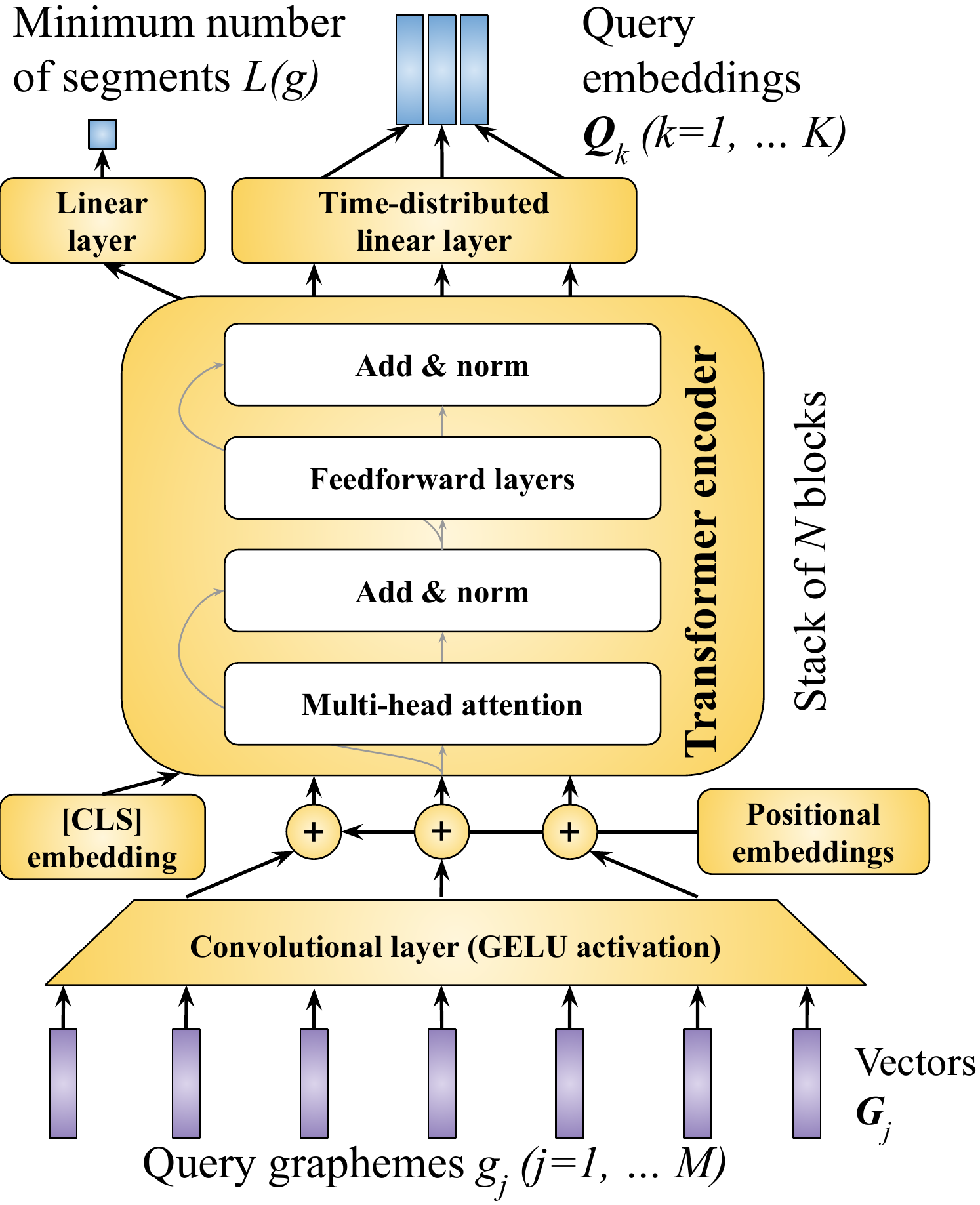}}

\caption{Architecture of the query encoder.}
\label{fig:query-encoder}
\vspace{-1em}

\end{figure}

To estimate the minimum number of segments $L(g)$, we add an extra token called \emph{[CLS]} to the input of the query encoder. A similar approach can be found in BERT pre-training where the \emph{CLS} token is used for the next-sentence-prediction task \cite{devlin2018bert}. The corresponding encoder output is transformed using a linear fully-connected layer to a scalar value $L(g)$. The training loss function of this network output is standard MSE (Fig. \ref{fig:query-encoder}). 

While the ability to condition the outputs on very distant parts of the input is one of the strongest features of the Transformer model, in STD, the occurrence of the putative hit depends only on local input features and distant dependencies are very rare. Therefore, we used \emph{attention masking} to limit the multi-headed attentions to attend only to a few neighboring input segments. Although attention masking is a common mechanism to introduce causality into Transformer models \cite{luo2022decbert}, we use it to introduce locality. The masking is implemented using the masking matrix containing ones on the diagonal and on a fixed number of super- and sub-diagonals. 

Another design choice that substantially reduces the number of trainable parameters in the networks is \emph{sharing the Transformer encoder} between the recognition output processing pipeline and the query pipeline. We propose to share only the parameters of the Transformer. Each pipeline still has its instance of the positional embeddings, convolution, and upsampling layers.

While the shared Transformers suggest a slightly worse performance in comparison with the separated Transformers architecture (Tab. \ref{tab:results}), we observed that it is beneficial for at least two reasons: (1) it stabilizes the training process, which always converges to a meaningful optimum and (2) it improves the performance in the \emph{multilingual setup}. As the multilingual setup, we call the model trained from the mix of data in two (or more) languages. We did not use any language identifier indicating the language of the query or the recognized hypothesis. The shared Transformers performed significantly better in the multilingual setup than the separated Transformers.

\subsection{Simplifications over Deep LSTM}

The proposed Transformer-based encoder$\cdot$encoder architecture brings some advantages over the baseline Deep LSTM. The Transformer can compensate for inaccurate time alignment of the CTC output and therefore we did not use the output masking as presented in \cite{svec22_interspeech}. In addition, the estimation of the minimum number of segments $L(g)$ for a given query is performed as a part of query embedding computation without requiring a separate model. And finally, the shared Transformer architecture and the multilingual setup lead to a substantially reduced number of trainable parameters when compared with Deep LSTM.

\subsection{Transformer block}
\label{sec:block}

We used a classical Transformer block as presented in \cite{Vaswani2017} with GELU activation functions in feed-forward layers. We used hyperparameter settings similar to the BERT models, particularly the BERT-Mini architecture \cite{bert_github}. BERT-Mini has a stack of four Transformer blocks with four self-attention heads, the dimensionality of embedding vectors is 256, and the dimensionality of feed-forward layers is 1024. The dropout probability used was 0.15.

\section{Dataset \& model description}
\label{sec:dataset}

The presented method was evaluated on the data from a USC-SFI MALACH archive in two languages -- English \cite{MALACHen} and Czech \cite{MALACHcz}. The training data for NN-based STD were extracted by blindly recognizing the archive using the DNN-HMM hybrid ASR with acoustic and language models trained from the manually transcribed part of the archives. Basic statistics are summarized in Tab. \ref{tab:stats} (more details are provided in \cite{svec2017}).

\begin{table}[b]
\vspace{-2em}
  \caption{Statistics of development and test sets \cite{svec2017}. ASR means DNN-HMM hybrid ASR.}
  \label{tab:stats}
  \centering
  \begin{tabular}{lrrrr}
    \toprule[0.9pt]
    & \multicolumn{2}{c}{English} & \multicolumn{2}{c}{Czech} \\
    \cmidrule(l){2-3} \cmidrule(l){4-5}
    & Dev & Test & Dev & Test \\
    \cmidrule(l){1-5}
    ASR vocabulary size & \multicolumn{2}{c}{243,699} & \multicolumn{2}{c}{252,082} \\
    \#speakers  & 10 & 10 & 10 & 10  \\
    OOV rate & 0.5\% & 3.2\% & 0.3\% & 2.6\%  \\
    ASR word error rate & 24.10 & 19.66 & 23.98 & 19.11 \\
    \#IV terms & 597 & 601 & 1680 & 1673 \\
    \#OOV terms & 31 & 6& 1145 & 948 \\
    dataset length $[$hours$]$ & 11.1 & 11.3 & 20.4 & 19.4 \\
    \bottomrule[0.9pt]
  \end{tabular}
\end{table}

We used the Wav2Vec model fine-tuned on the MALACH data to generate the grapheme confusion networks. For English, we started with publicly available pre-trained Wav2Vec 2.0 Base model~\cite{wav2vec_base}. For Czech, we used the ClTRUS model \cite{lehecka22_interspeech} pre-trained from more than 80 thousand hours of Czech speech data following the same pretraining steps as for the base Wav2Vec 2.0 model \cite{Baevski2020}. 

The Fairseq tool \cite{ott2019fairseq} was used for fine-tuning. We sliced long training audio signals on speech pauses not to exceed the length of 30\,s. We removed non-speech events and punctuation from the transcripts and mapped all graphemes into lowercase. The pre-trained models were fine-tuned for 80k updates with a peak learning rate of $8 \times 10^{-5}$ for English and $2 \times 10^{-5}$ for Czech, respectively. The CTC classification layer predicts probabilities of 53 symbols for English and 51 for Czech, and the output frame length is 0.02s for both models.

It is important to note, that the CTC loss does not guarantee the precise time alignment of the generated sequence of symbols. However, the timing produced by the fine-tuned models is sufficient to perform STD over the generated hypothesis as was shown for {in-vocabulary} terms in \cite{svec22_interspeech}.

The in-vocabulary (IV) and out-of-vocabulary (OOV) terms were selected automatically from the development and test data based on the DNN-HMM recognition vocabulary. We filtered all possible terms so that the terms are not substrings of other terms in the dataset nor the words in the vocabulary. The numbers reported in this paper are directly comparable to results presented in \cite{svec2017,svec21_interspeech,svec22_interspeech}.


\section{Experimental results}
\label{sec:experimental}

For training the Transformer-based encoder$\cdot$encoder architecture, we used ADAM optimizer with a learning rate warm-up. The warm-up raised the learning rate from 0 to $10^{-4}$ in the initial 80k training steps and then the learning rate decayed linearly to 0 in the next 720k training steps. We used lower learning rates and longer training than Deep LSTM because the training of Transformers tends to collapse if higher rates are used. The models were trained using $N=256$ (number of confusion network segments in the input) and $M=16$ (maximum length of the query).

In the experiments, we first optimized the MTWV metric \cite{wegmann2013} on the development dataset (Tab. \ref{tab:results}). Then, using a given architecture, the optimal decision threshold was determined and applied to the test dataset and the ATWV metric was computed (Tab. \ref{tab:test}). In the experiments, we used both the in- and out-of-vocabulary terms. The presented method is designed to generalize from seen IV terms to detect the OOV terms.

Tab. \ref{tab:results} follows the changes proposed in Sec. \ref{sec:transformer}. The drop-in replacement of Deep LSTM with the vanilla Transformer degraded the performance. The addition of convolutional and upsampling layers led to a minor improvement. In the experiments, we searched for an optimum 1D convolution width and stride with a grid search. We found that width 3 and stride 2 maximized the MTWV, which leads to $K=8$ query embeddings per each query.

In the next step, we added attention masking. Again, we swept across an interval of different widths of the attention mask, and finally, we used the diagonal matrix of ones with two super- and two sub-diagonals. In other words, the Transformer blocks attend to the current, two preceding, and two following time steps. We have to mention that this does not imply the context is just five segments because several Transformer blocks are stacked; therefore, the context of the last layer is wider than these five segments.

Then, we tied the parameters of the Transformers used in the hypothesis and query encoders to effectively reduce the number of trainable parameters from 7.3M to 4.2M (cf. Deep LSTMs 6.9M). We observed an interesting behavior of the shared Transformer -- if it is trained for each of the languages (English and Czech) separately, the MTWV is slightly worse. Considering the multilingual setup, where both English and Czech training data are mixed and the resulting single network is evaluated separately on English and Czech, the difference between the shared and separated Transformer is exactly the opposite. We also observed that using the shared Transformer stabilized the training process, which never collapsed as in the case of the separated Transformer, where it rarely occurred. 

\begin{table}[t]
    \small
  \caption{Results on the development dataset (MTWV$\uparrow$).}
  \label{tab:results}
  \centering
  \begin{tabular}{p{5cm}lrr}
    \cmidrule[0.9pt](l){1-3}
     & English & Czech \\
    \cmidrule(l){1-3}
     Deep LSTM (baseline \cite{svec22_interspeech}) & 0.8308 & 0.8987 \\
    \cmidrule(l){1-3}
     \multicolumn{3}{l}{\textbf{Transformer, monolingual setup} (proposed method)} \\    
        Vanilla Transformer & 0.8163 & 0.8808 \\
        + Convolution layers & 0.8395 & 0.8905 \\
        + Attention masking & 0.8588 & 0.9261 \\
        + Shared parameters & 0.8545 & 0.9144 \\
    \cmidrule(l){1-3}
     \multicolumn{3}{l}{\textbf{Transformer, multilingual setup} (proposed method)} \\    
        Separated Transformers & 0.8196 & 0.8922 \\
        Shared Transformers    & 0.8593 & 0.9229 \\
    \cmidrule[0.9pt](l){1-3}
  \end{tabular}
  \caption{Results on the test dataset (ATWV$\uparrow$).}
  \label{tab:test}
  \centering
  \begin{tabular}{p{5cm}lrr}
    \cmidrule[0.9pt](l){1-3}
     & English & Czech \\
    \cmidrule(l){1-3}
     Deep LSTM (baseline \cite{svec22_interspeech}) & 0.7616 & 0.9100 \\
     \textbf{Transformer, monolingual setup}        & 0.7938 & 0.9120 \\
     \textbf{Transformer, multilingual setup}       & 0.7925 & 0.9062 \\
    \cmidrule[0.9pt](l){1-3}
  \end{tabular}
  \vspace{-1em}
\end{table}

As the final step, the presented architectures were evaluated on the test dataset (Tab. \ref{tab:test}). The ATWV on English data follows the improvement in MTWV on the development dataset. For Czech data, we achieved similar performance as the baseline, probably because the STD reached its limits on this dataset.

\section{Conclusion}
\label{sec:conclusion}

We proposed an NN-based STD method employing two BERT-like encoders. We modify the vanilla Transformer by adding convolutional and upsampling layers. For the hypothesis encoder, we also used the attention masking mechanism.
The presented modifications of the NN-based STD employing the Transformer encoder$\cdot$encoder architecture achieved a 0.03 improvement in MTWV/ATWV on development and test datasets except for the Czech test dataset, while reducing the number of trainable parameters.

The proposed modifications are usable not only in STD tasks employing graphemic queries and recognized hypotheses but also in related tasks such as QbE, KWS, or WWD. The ability of the Transformer encoder$\cdot$encoder model to share parameters between the hypothesis encoder and query encoder and between different input languages opens further research questions, such as the possibility of combining STD (graphemic query) and QbE (spoken query) in a single, multi-task trained model. This approach can focus on directly using the Wav2Vec posterior probabilities without the conversion to grapheme confusion networks.

\section{Acknowledgement}

Computational resources were supplied by the project "e-Infrastruktura CZ" (e-INFRA CZ LM2018140 ) supported by the Ministry of Education, Youth, and Sports of the Czech Republic.


\bibliographystyle{IEEEbib}
\bibliography{refs}

\end{document}